\documentclass{article}
\usepackage{algorithm}
\usepackage{algpseudocode}
\usepackage{amsmath}
\usepackage{booktabs}
\usepackage{graphicx}
\usepackage{hyperref}
\usepackage{multirow}
\usepackage{xcolor}
\usepackage[preprint]{spconf}
\usepackage{caption}
\newcommand{\sd}[1]{$_{\pm #1}$}
\title{StreamHear: Domain-Adapted Pseudo-Labeling for Semi-Supervised Streaming Speech Recognition}
\name{Zefang Liu, Chenyang Zhu, Sangwoo Cho, Xujun Peng, Shi-Xiong Zhang, Sambit Sahu}
\address{Capital One, USA\\zefang.liu@capitalone.com}
\begin{document}
\ninept
\maketitle
\begin{abstract}
Streaming automatic speech recognition (ASR) underperforms on domain-shifted target audio, where labeled in-domain data is costly to prepare while unlabeled audio is abundant. We present StreamHear, a semi-supervised pipeline that adapts a pretrained streaming student by fine-tuning an offline transducer teacher on the labeled training set, generating pseudo-labels on the unlabeled portion, and fine-tuning the student on the mixture. We further introduce a prior-regularized dynamic-programming realignment step that fixes chunk-level word placement using an ASR-hypothesis anchor. Across four datasets spanning financial calls, prepared read speech, and phone-quality dialogue, StreamHear consistently outperforms supervised student fine-tuning and narrows the gap to the offline teacher.
\end{abstract}
\section{Introduction}

Pretrained streaming automatic speech recognition (ASR) models, such as cache-aware FastConformer-RNN-T~\cite{noroozi2024stateful}, enable low-latency transcription for real-time applications. Yet on domain-shifted target audio such as financial earnings calls, international English, and phone-quality customer-service conversation, they still underperform, struggling with entity-dense vocabulary, accented pronunciations, and narrowband acoustics. In-domain fine-tuning is the natural remedy, but labeled in-domain audio is costly to prepare at chunk granularity while unlabeled in-domain audio is abundant.

Semi-supervised pseudo-labeling addresses this gap and has been widely studied in ASR, from foundational iterative self-training~\cite{kahn2020self,xu2020iterative,likhomanenko2021slimipl} to teacher-student momentum methods~\cite{higuchi2021momentum}, dynamic pseudo-label caches with checkpoint-averaged teachers~\cite{tadevosyan2025unified}, and correctors based on audio-aware large language models (LLMs)~\cite{prakash2025better,liu2026rehear}. Despite varied designs, these methods share a common axis: iterative refinement, exponential moving average (EMA) teachers, or auxiliary neural machinery to raise pseudo-label quality. For cache-aware streaming ASR, however, a much simpler recipe using a fixed, domain-adapted offline teacher has not been reported.

In this work, we present \textbf{StreamHear}, a semi-supervised pipeline for adapting cache-aware streaming ASR to a target domain, with three contributions. First, the \textbf{StreamHear pipeline} fine-tunes an offline transducer teacher on the labeled in-domain training set, generates pseudo-labels on the unlabeled portion, and fine-tunes the streaming student on the mixture, all in a single pass without iteration or auxiliary machinery. Second, we introduce a prior-regularized \textbf{dynamic-programming realignment} step that redistributes ground-truth words across chunk boundaries in the training data using an ASR-hypothesis anchor, correcting residual placement errors left by Connectionist Temporal Classification (CTC) Segmentation~\cite{kurzinger2020ctc}. Third, our \textbf{empirical study on four datasets} shows that StreamHear consistently outperforms supervised student fine-tuning and narrows the gap to the offline teacher, with ablations confirming robustness across latency and context configurations.

\section{Related Work}

Pseudo-labeling for ASR originated with early self-training work~\cite{kahn2020self,chen2020semi} and was formalized into iterative pipelines by IPL~\cite{xu2020iterative} and slimIPL~\cite{likhomanenko2021slimipl}. Subsequent methods have raised label quality with EMA teachers that continuously trail the student (MPL~\cite{higuchi2021momentum}), dynamic pseudo-label caches with checkpoint-averaged teachers~\cite{tadevosyan2025unified}, confidence- and uncertainty-based filtering~\cite{jin2022filter,khurana2021unsupervised,kim2025uncertainty}, incremental retraining with data filtering~\cite{carofilis2025better}, and audio-aware LLM correctors that condition on source audio~\cite{prakash2025better,liu2026rehear}. The closest prior work applies pseudo-labeling to streaming students: iterative noisy-student training with a large offline teacher paired with a small streaming student~\cite{hwang2022pseudo}, and knowledge distillation from Whisper into streaming Transformer-Transducers trained from scratch~\cite{thorbecke2024fast}. Neither targets cache-aware streaming~\cite{noroozi2024stateful}, an architecture that passes activation caches between chunks and eliminates the training-inference gap. Across these methods, the teacher is either updated during training or augmented with auxiliary neural machinery, and the streaming setting is served by iterative noisy-student training or from-scratch distillation. Departing from these paradigms, StreamHear uses a fixed, domain-adapted offline transducer as its teacher, mixing a single round of its pseudo-labels with the labeled training set for one student fine-tuning pass, thereby adapting a pretrained cache-aware streaming transducer without iteration or auxiliary machinery.

\section{Methodology}

Our proposed \textbf{StreamHear} framework consists of three sequential learning stages preceded by a chunk-level data preparation step. We denote the offline teacher ASR model as $M_T$ and the cache-aware streaming student ASR model as $M_S$. Given a labeled in-domain training set $D_L = \{(x_l, y_l)\}$ and an unlabeled in-domain set $D_U = \{x_u\}$, the pipeline executes teacher fine-tuning, pseudo-label generation, and student fine-tuning in order, as summarized in Algorithm~\ref{alg:streamhear}. Unlike iterative pseudo-labeling frameworks, StreamHear runs each stage exactly once and yields a single deployable streaming checkpoint.

\begin{algorithm}[h]
\caption{StreamHear: Pseudo-Labeling for Streaming ASR}
\label{alg:streamhear}
\begin{algorithmic}[1]

\Statex \textbf{Models:} Offline teacher $M_T$, Streaming student $M_S$
\Statex \textbf{Data:} Labeled $D_L = \{(x_l, y_l)\}$, Unlabeled $D_U = \{x_u\}$

\vspace{0.1cm}
\State \textcolor{gray}{\textit{// Teacher fine-tuning}}
\State Fine-tune $M_T$ on $D_L$

\State \textcolor{gray}{\textit{// Pseudo-label generation}}
\State $D'_U \leftarrow \{ (x_u, M_T(x_u)) \mid x_u \in D_U \}$

\State \textcolor{gray}{\textit{// Optional confidence filter}}
\State Retain the top-$K$\% of $D'_U$ ranked by teacher log-likelihood

\State \textcolor{gray}{\textit{// Student fine-tuning}}
\State Fine-tune $M_S$ on $D_L \cup D'_U$

\State \textbf{return} $M_S$

\end{algorithmic}
\end{algorithm}

\textbf{Teacher fine-tuning.} The offline teacher $M_T$, a full-context transducer, is fine-tuned on the labeled training set $D_L$ under the standard transducer loss. This stage domain-adapts the teacher to the target audio distribution, raising the quality of the pseudo-labels it subsequently generates. \textbf{Pseudo-label generation.} The fine-tuned teacher transcribes the unlabeled portion $D_U$ once with greedy decoding, yielding pseudo-labeled chunks $D'_U = \{(x_u, y'_u)\}$ where $y'_u = M_T(x_u)$. Optionally, pseudo-labels can be filtered to the top-$K$\% ranked by sequence-average teacher log-likelihood. \textbf{Student fine-tuning.} The cache-aware streaming student $M_S$ is fine-tuned on the mixed manifest $D_L \cup D'_U$ under the standard RNN-T loss.

\textbf{Chunk-level data preparation.} Cache-aware streaming training requires each chunk to carry only the words spoken within its time interval. We segment full-length recordings using voice-activity detection (VAD) and compute word-level timestamps by applying CTC-Segmentation~\cite{kurzinger2020ctc} on top of a self-supervised acoustic model. This baseline leaves residual placement drift near non-speech tags and VAD boundaries. To repair these errors, we introduce a prior-regularized dynamic-programming (DP) realignment step, summarized in Algorithm~\ref{alg:realignment}. We flatten the ground-truth transcript and per-chunk ASR hypothesis into two word streams, each word tagged with its host chunk index. We then run Needleman-Wunsch alignment scored by word match rewards, skip costs, and a prior penalty on the chunk-index displacement between paired words. Matched ground-truth words inherit their hypothesis partner's chunk; unmatched words keep their CTC-Segmentation chunk, and a two-pass sweep clamps them into monotonic order.

\begin{algorithm}[h]
\caption{Dynamic-Programming Realignment}
\label{alg:realignment}
\begin{algorithmic}[1]

\Statex \textbf{Input:} Per-chunk ground-truth text and ASR hypothesis, match reward $r$, mismatch cost $c_m$, skip costs $c_g, c_h$, prior weight $\lambda$
\Statex \textbf{Output:} Updated word-to-chunk assignment

\vspace{0.1cm}
\State Concatenate ground-truth words into $(w_1, \dots, w_N)$ and hypothesis words into $(h_1, \dots, h_M)$, tagging each with its host chunk index $c^{\text{gt}}_i$, $c^{\text{hyp}}_j$
\State Initialize $dp[0,0] = 0$, $dp[i,0] = -i \cdot c_g$ for $i = 1..N$, $dp[0,j] = -j \cdot c_h$ for $j = 1..M$
\State For $i = 1..N$ and $j = 1..M$, set $dp[i, j]$ to the best of: diagonal $dp[i{-}1, j{-}1] + s_{ij} - \lambda |c^{\text{hyp}}_j - c^{\text{gt}}_i|$ with $s_{ij}{=}r$ if $w_i {\approx} h_j$ else $-c_m$; vertical $dp[i{-}1, j] - c_g$; horizontal $dp[i, j{-}1] - c_h$
\State Traceback from $(N, M)$: matched $w_i$ inherits $c^{\text{hyp}}_j$; unmatched $w_i$ keeps $c^{\text{gt}}_i$
\State Two-pass sweep clamps unmatched words into monotonic chunk order

\end{algorithmic}
\end{algorithm}

\section{Experiments}

This section describes the datasets and their chunk-level preparation, the training and inference setup, the main WER comparison against pretrained and fine-tuned baselines, and five ablations probing alignment correction, pseudo-label pool scaling, context sensitivity, per-latency retraining, and the streaming student architecture.

\subsection{Datasets}

We evaluate on three public English corpora plus one proprietary call-center dataset. \textbf{Earnings-21}~\cite{del2021earnings} contains 44 quarterly earnings calls across nine financial sectors with entity-dense transcripts. \textbf{Earnings-22}~\cite{del2022earnings} contains 125 English earnings calls sourced from global companies spanning seven world-English regions; four files with predominantly non-English audio are excluded at prepare time, leaving 121 effective files. \textbf{SPGISpeech}~\cite{oneill2021spgispeech} is a large-scale corpus of professionally-transcribed financial teleconference audio (mix of narrated presentations and spontaneous Q\&A); we use its publicly-released small subset, further downsampled per~\cite{liu2026rehear} ratios (10\% labeled train, 30\% unlabeled, 10\% validation, 10\% test) at the session level. \textbf{BankCall} is a proprietary human-annotated set of 155 stereo banking customer-service calls (channel 1 customer, channel 2 agent); the unlabeled pool is 275 additional calls drawn as a 30\% subsample of separately-collected raw audio from the same domain. Table~\ref{tab:datasets} lists per-split hour totals.

\begin{table}[!t]
\centering
\caption{Detailed statistics of the datasets utilized in experiments, reporting the total duration in hours for each partition split and the average utterance length in seconds.}
\label{tab:datasets}
\small
\resizebox{\linewidth}{!}{%
\begin{tabular}{lrrrr}
\toprule
\textbf{Split} & \textbf{Earnings-21} & \textbf{Earnings-22} & \textbf{SPGISpeech} & \textbf{BankCall} \\
\midrule
Labeled Train & 10.72 & 24.76 & 19.58 & 12.96 \\
Labeled Val & 3.09 & 9.16 & 10.66 & 2.62 \\
Labeled Test & 4.88 & 12.38 & 10.23 & 2.88 \\
Unlabeled & 18.85 & 63.01 & 58.63 & 44.33 \\
\midrule
\textit{Total} & 37.54 & 109.31 & 99.10 & 62.79 \\
\textit{Utterance (s)} & 23.3 & 23.0 & 9.2 & 6.2 \\
\bottomrule
\end{tabular}%
}
\end{table}

\begin{table*}[t]
\centering
\caption{Word Error Rate (WER, \%) comparison across all datasets. We report the mean and standard deviation (subscript) over $5$ seeds on the labeled test and unlabeled splits. BankCall reports labeled test only (unlabeled pool has no transcripts). \textbf{Bold} indicates the best streaming-student result per column.}
\label{tab:main}
\small
\begin{tabular}{lccccccc}
\toprule
\multirow{2.5}{*}{\textbf{Method}} & \multicolumn{2}{c}{\textbf{Earnings-21}} & \multicolumn{2}{c}{\textbf{Earnings-22}} & \multicolumn{2}{c}{\textbf{SPGISpeech}} & \textbf{BankCall} \\
\cmidrule(lr){2-3} \cmidrule(lr){4-5} \cmidrule(lr){6-7} \cmidrule(lr){8-8}
& \textbf{Test} & \textbf{Unlabeled} & \textbf{Test} & \textbf{Unlabeled} & \textbf{Test} & \textbf{Unlabeled} & \textbf{Test} \\
\midrule
Teacher & 6.51\sd{0.00} & \phantom{0}8.35\sd{0.00} & 12.25\sd{0.00} & 10.62\sd{0.00} & 3.97\sd{0.00} & 4.02\sd{0.00} & 12.71\sd{0.00} \\
FT Teacher & 6.03\sd{0.05} & \phantom{0}7.68\sd{0.03} & 11.24\sd{0.03} & \phantom{0}9.55\sd{0.01} & 2.20\sd{0.02} & 2.15\sd{0.01} & \phantom{0}8.85\sd{0.11} \\
\midrule
Student & 8.27\sd{0.00} & 10.36\sd{0.00} & 13.55\sd{0.00} & 12.31\sd{0.00} & 3.27\sd{0.00} & 3.18\sd{0.00} & 13.60\sd{0.00} \\
FT Student & 7.19\sd{0.09} & \phantom{0}9.62\sd{0.10} & 12.46\sd{0.06} & 10.81\sd{0.01} & 2.77\sd{0.02} & 2.70\sd{0.03} & 10.68\sd{0.13} \\
StreamHear & \textbf{6.63}\sd{0.04} & \textbf{\phantom{0}7.77}\sd{0.03} & \textbf{11.76}\sd{0.01} & \textbf{\phantom{0}9.56}\sd{0.01} & \textbf{2.59}\sd{0.03} & \textbf{2.26}\sd{0.01} & \textbf{\phantom{0}9.80}\sd{0.05} \\
\bottomrule
\end{tabular}
\end{table*}

For long-form Earnings-21 and Earnings-22, we standardize raw recordings to 16 kHz, normalize non-speech annotations to unified tags, and produce chunk-level training data via VAD segmentation with pyannote-3.0~\cite{bredin2020pyannote}, CTC-Segmentation forced alignment using Parakeet-CTC-0.6B~\cite{rekesh2023fast}, and the DP realignment step of Algorithm~\ref{alg:realignment} with a Whisper-Large-v3~\cite{radford2023robust} hypothesis anchor. VAD keeps each chunk within Whisper's 30 s context so the anchor covers it in full. Longer audio can be anchored with an audio-LLM such as Voxtral-Mini-3B~\cite{liu2025voxtral} that batches multi-window inputs across a longer context. SPGISpeech utterances arrive pre-chunked from the source parquets, so preparation reduces to session-level partitioning at ReHear~\cite{liu2026rehear} ratios. For BankCall, human annotations already provide chunk-level segments with speaker channel and timestamps, so no VAD or forced alignment is applied on the labeled portion; digit-mask redactions are filled by a strict-majority vote among three off-the-shelf ASR models (Whisper-Large-v3, Voxtral-Mini-3B, Parakeet-TDT-0.6B-v3) plus a human review pass, and the labeled split is stratified by call reason. The unlabeled BankCall portion is segmented via VAD only. Across all datasets, chunks with empty transcripts are dropped, and partitioning is enforced at the source file or session level.

\subsection{Experimental Setup}

\textbf{Models.} The offline teacher is Parakeet-TDT-0.6B-v3~\cite{rekesh2023fast}, a 0.6B-parameter FastConformer transducer with token-and-duration decoding. The primary streaming student is Nemotron-Speech-Streaming-EN-0.6B~\cite{noroozi2024stateful}, a 0.6B cache-aware FastConformer-RNN-T trained with multi-latency right-context (RC) sampling; unless noted, we run inference at $\text{RC}{=}1$ ($80$ ms chunk plus $80$ ms right-context lookahead, for $160$ ms of algorithmic latency). The student-choice ablation additionally evaluates Nemotron-3.5-ASR-Streaming-0.6B, a multilingual variant with $\text{LC}{=}56$ and English language-ID conditioning.

\textbf{Training.} All fine-tuning runs share the same recipe: AdamW optimizer with learning rate $2\times 10^{-4}$, betas $[0.9, 0.98]$, weight decay $10^{-3}$; cosine schedule with $10\%$ warmup and minimum learning rate $10^{-6}$; $10$ epochs; bf16 precision; SpecAugment ($2$ freq masks $\times$ $10$ time masks). Effective batch size $64$ ($4$ A100-SXM-40GB GPUs, per-GPU batch $4$, gradient accumulation $4$) for the streaming student; teacher fine-tuning uses per-GPU batch $2$ with gradient accumulation $8$ to fit the RNN-T joint on Earnings-length chunks; Nemotron-3.5 drops to per-GPU batch $1$ with gradient accumulation $16$. Each experimental cell is repeated for $5$ seeds with matched teacher-student pairings for the pseudo-labeling runs; we report mean and standard deviation. All training and inference use the NVIDIA NeMo Speech framework\footnote{\url{https://github.com/NVIDIA-NeMo/Speech}}~\cite{kuchaiev2019nemo}.

\textbf{Inference and metrics.} Inference uses greedy decoding under the cache-aware streaming attention mask. Word error rate (WER) is computed after a text normalization pipeline: bracket removal, number and typography normalization, lowercasing, contraction expansion, spelling and compound normalization, diacritic removal, filler-word removal, and punctuation removal. Latency is reported as the algorithmic latency ($(1{+}\text{RC})\times 80$ ms) imposed by the right-context lookahead.

\subsection{Experimental Results}

As presented in Table~\ref{tab:main}, StreamHear consistently outperforms the fine-tuned student on all four datasets: by $0.18$ to $0.88$ percentage points (pp) of word error rate on the labeled test split, and by $0.44$ to $1.85$ pp on the held-out unlabeled split of the three public benchmarks. Against the offline fine-tuned teacher, StreamHear narrows the labeled-test gap to at most $0.95$ pp and the unlabeled gap to at most $0.11$ pp; on Earnings-22 unlabeled, the streaming student essentially matches the offline teacher ($9.56\%$ vs $9.55\%$). These results confirm that pseudo-labels from a domain-adapted offline teacher transfer to a streaming student in a single fine-tuning pass, without iterative refinement or auxiliary neural machinery. On BankCall's stereo audio, StreamHear's $0.88$ pp gain is essentially symmetric across the customer and agent channels, bringing streaming WER to $12.55\%$ and $7.00\%$ respectively ($-0.90$ and $-0.85$ pp vs the fine-tuned student) and preserving the intrinsic ${\sim}5.5$ pp customer/agent WER gap that reflects acoustic quality rather than a labeling asymmetry.

\subsection{Ablation Studies}

We report five ablations. All experiments use Nemotron-Speech-Streaming-EN-0.6B on Earnings-21 unless noted; all numbers are mean WER (\%) over $5$ seeds under the same normalization pipeline as Table~\ref{tab:main}.

\subsubsection{Alignment Correction}

We measure alignment quality by running the pretrained offline teacher (Parakeet-TDT-0.6B-v3) on each dataset's chunk text before and after DP realignment, computing WER against the teacher's own audio hypothesis on the same chunks. Since the DP step uses Whisper-Large-v3 as its position prior (not Parakeet), the teacher is an independent anchor. Realignment cuts alignment-quality WER by roughly $9$ pp on both datasets (Table~\ref{tab:alignment}) while changing chunk count by at most $0.4\%$, confirming that words are reassigned to the correct chunks rather than merely dropped.

\begin{table}[h]
\centering
\caption{Alignment-quality WER (\%) on chunk text against a pretrained Parakeet-TDT-0.6B-v3 audio hypothesis, before and after DP realignment.}
\label{tab:alignment}
\small
\begin{tabular}{lccc}
\toprule
\textbf{Dataset} & \textbf{Aligned} & \textbf{Realigned} & \textbf{Change} \\
\midrule
Earnings-21 & 17.46 & \phantom{0}8.01 & $-9.45$ \\
Earnings-22 & 20.54 & 10.97 & $-9.57$ \\
\bottomrule
\end{tabular}
\end{table}

\subsubsection{Confidence Filtering}

We sweep a top-$K$\% confidence filter on pseudo-labels ranked by sequence-average teacher log-likelihood, for $K{\in}\{25, 50, 75, 100\}$, on Earnings-21 and Earnings-22 (Table~\ref{tab:pseudo-label-scaling}). Larger $K$ monotonically improves WER on both datasets and both splits; discarding low-confidence pseudo-labels never helps. Even $K{=}25\%$ already outperforms the fine-tuned student, indicating that a domain-adapted teacher produces uniformly usable pseudo-labels and that quantity outweighs top-$K$ quality.

\begin{table}[h]
\centering
\caption{Pseudo-label pool scaling with a top-$K$\% confidence filter. WER (\%, mean$_{\pm\text{std}}$) as labeled test / unlabeled.}
\label{tab:pseudo-label-scaling}
\small
\begin{tabular}{lcc}
\toprule
\textbf{Top-K} & \textbf{Earnings-21} & \textbf{Earnings-22} \\
\midrule
25\% & 6.92\sd{0.03} / 8.65\sd{0.07} & 11.99\sd{0.03} / 10.17\sd{0.02} \\
50\% & 6.83\sd{0.06} / 8.28\sd{0.05} & 11.86\sd{0.02} / \phantom{0}9.87\sd{0.03} \\
75\% & 6.72\sd{0.08} / 8.01\sd{0.02} & 11.80\sd{0.04} / \phantom{0}9.68\sd{0.03} \\
100\% & \textbf{6.63\sd{0.04} / 7.77\sd{0.03}} & \textbf{11.76\sd{0.01} / \phantom{0}9.56\sd{0.01}} \\
\bottomrule
\end{tabular}
\end{table}

\subsubsection{Context Sweeps}

\begin{figure}[h]
\centering
\includegraphics[width=0.8\linewidth]{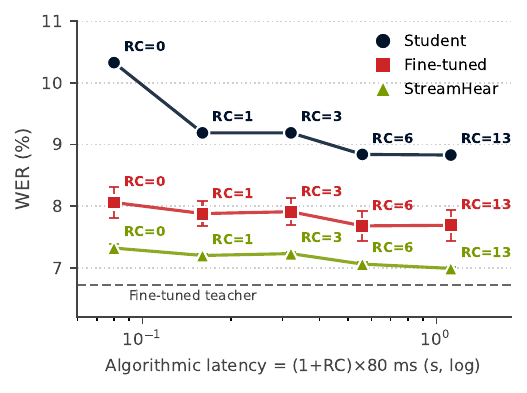}
\caption{Right-context (RC) sweep on Earnings-21 labeled test WER (\%) at fixed $\text{LC}{=}70$. Algorithmic latency is $(1{+}\text{RC})\times 80$ ms. Error bars are ${\pm}1$ std; student inference is deterministic, and StreamHear std (${\leq}0.06$ pp) is smaller than the marker.}
\label{fig:right-context}
\end{figure}

\begin{figure}[h]
\centering
\includegraphics[width=0.8\linewidth]{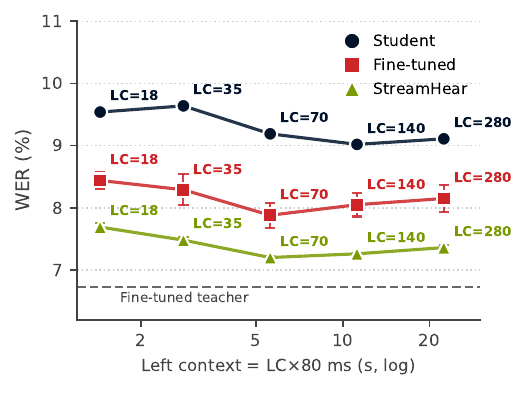}
\caption{Left-context (LC) sweep on Earnings-21 labeled test WER (\%) at fixed $\text{RC}{=}1$. Left context is $\text{LC}\times 80$ ms. Error bars are ${\pm}1$ std; StreamHear std (${\leq}0.05$ pp) is smaller than the marker.}
\label{fig:left-context}
\end{figure}

We decode the base checkpoint, the fine-tuned student, and StreamHear across right-context $\text{RC}{\in}\{0, 1, 3, 6, 13\}$ at fixed $\text{LC}{=}70$, giving algorithmic latencies from $80$ ms to $1.12$ s. The streaming student is pretrained at RCs $\{0, 1, 6, 13\}$; $\text{RC}{=}3$ is an interior point reached only through multi-latency generalization. The fine-tuned student and StreamHear are each trained once at the default configuration $[\text{LC}{=}70, \text{RC}{=}1]$ and then decoded at every RC in the sweep. Figure~\ref{fig:right-context} plots labeled test WER (\%) against the algorithmic latency $(1{+}\text{RC})\times 80$ ms for the three model conditions. Under all three, WER decreases with RC and saturates by $\text{RC}{=}6$. StreamHear preserves the qualitative shape of the base and fine-tuned curves, so switching from $\text{RC}{=}1$ to $\text{RC}{=}13$ yields only $0.20$ pp at the cost of $0.96$ s of extra algorithmic latency. We similarly sweep left-context $\text{LC}{\in}\{18, 35, 70, 140, 280\}$ at fixed $\text{RC}{=}1$ (Figure~\ref{fig:left-context}), decoding the same checkpoints at each LC; all three curves reach their minima near the trained value $\text{LC}{=}70$ and rise on either side, showing a much weaker effect than RC (spread ${\leq}0.65$ pp across the whole curve).

\subsubsection{Per-Latency Retraining}

We fine-tune a separate checkpoint at each $\text{RC}{\in}\{0, 3, 6, 13\}$ (both supervised and StreamHear) and evaluate at the matching RC. Compared against decoding the single checkpoint trained at $\text{RC}{=}1$ at those RCs (Figure~\ref{fig:right-context}), supervised per-RC training outperforms decode-sweep by $0.02$ to $0.06$ pp on labeled test across the four RCs, while StreamHear per-RC training matches decode-sweep to within $0.01$ pp at every RC. A single StreamHear checkpoint trained at $\text{RC}{=}1$ therefore suffices for the whole latency curve, obviating per-RC retraining.

\subsubsection{Student Choice}

We replace the primary student, hereafter Nemotron-EN (Nemotron-Speech-Streaming-EN-0.6B), with the multilingual Nemotron-ML (Nemotron-3.5-ASR-Streaming-0.6B, trained at $\text{LC}{=}56$, prompt-conditioned decoder) and re-run the full StreamHear pipeline on Earnings-21 (Table~\ref{tab:student-choice}). StreamHear outperforms supervised FT by $0.44$ pp on labeled test and $2.28$ pp on unlabeled, mirroring Nemotron-EN's $0.56$ and $1.85$ pp deltas: both students absorb a similar amount of extra signal from pseudo-labels. Absolute WER for Nemotron-ML stays $0.89$-$2.82$ pp higher than Nemotron-EN at every stage (Student, FT Student, StreamHear), so the choice of pretrained backbone sets a ceiling that StreamHear does not fully lift, but the recipe generalizes across streaming student architectures.

\begin{table}[h]
\centering
\caption{Student-choice ablation on Earnings-21 WER (\%, mean$_{\pm\text{std}}$) as labeled test / unlabeled.}
\label{tab:student-choice}
\small
\begin{tabular}{lcc}
\toprule
\textbf{Method} & \textbf{Nemotron-EN} & \textbf{Nemotron-ML} \\
\midrule
Student & \phantom{0}8.27\sd{0.00} / 10.36\sd{0.00} & 10.53\sd{0.00} / 13.18\sd{0.00} \\
FT Student & \phantom{0}7.19\sd{0.09} / \phantom{0}9.62\sd{0.10} & \phantom{0}8.54\sd{0.11} / 10.94\sd{0.13} \\
StreamHear & \textbf{\phantom{0}6.63\sd{0.04} / \phantom{0}7.77\sd{0.03}} & \textbf{\phantom{0}8.10\sd{0.07} / \phantom{0}8.66\sd{0.05}} \\
\bottomrule
\end{tabular}
\end{table}

\section{Conclusion}

We presented StreamHear, a semi-supervised recipe that adapts a cache-aware streaming ASR student to a target domain by fine-tuning on pseudo-labels from a domain-adapted offline teacher. A prior-regularized DP realignment step redistributes ground-truth words across chunk boundaries using an ASR hypothesis anchor. Across Earnings-21, Earnings-22, SPGISpeech, and a proprietary call-center corpus, BankCall, StreamHear consistently outperforms the supervised streaming student on both labeled test and unlabeled, and narrows the gap to the offline teacher. Ablations show these gains are robust across latency operating points, pseudo-label pool sizes, context settings, and student architectures.

\newpage
\bibliographystyle{IEEEbib}
\bibliography{refs}
\end{document}